\documentclass[10pt]{article}
\usepackage{amsmath}
\usepackage{times}
\usepackage{graphicx}
\usepackage{color}
\usepackage{multirow}

\usepackage[authoryear]{natbib}
\usepackage{rotating}
\usepackage{latexsym}
\usepackage{subfigure} 

\newcommand{\captionfonts}{\normalsize}

\makeatletter
\long\def\@makecaption#1#2{%
  \vskip\abovecaptionskip
  \sbox\@tempboxa{{\captionfonts #1: #2}}%
  \ifdim \wd\@tempboxa >\hsize
    {\captionfonts #1: #2\par}
  \else
    \hbox to\hsize{\hfil\box\@tempboxa\hfil}%
  \fi
  \vskip\belowcaptionskip}
\makeatother

\begin{document}
\hspace{13.9cm}1

\ \vspace{20mm}\\


{\LARGE Statistics of Visual Responses to Object Stimuli from Primate AIT Neurons to DNN Neurons}

\ \\
{\bf \large Qiulei Dong$^{\displaystyle 1, \displaystyle 2, \displaystyle 3}$, Zhanyi Hu$^{\displaystyle 1, \displaystyle 2, \displaystyle 3 }$}\\
{$^{\displaystyle 1}$National Laboratory of Pattern Recognition, Institute of Automation, Chinese Academy of Sciences, Beijing
100190, China.}\\
{$^{\displaystyle 2}$University of Chinese Academy of Sciences, Beijing 100049, China.}\\
{$^{\displaystyle 3}$CAS Center for Excellence in Brain Science and Intelligence Technology, Beijing 100190, China.}\\
%

{\bf Keywords:} Inferotemporal Cortex, Single-Neuron Selectivity, Population Sparseness, Intrinsic Dimensionality, Deep Neural Networks(DNNs)

\thispagestyle{empty}
\markboth{}{NC instructions}
\ \vspace{-0mm}\\

%
\begin{center} {\bf Abstract} \end{center}
Cadieu et al. \citep{Cadieu2014} reported that deep neural networks (DNNs) could rival the representation of primate inferotemporal cortex for core visual object recognition. Lehky et al. \citep{Lehky2011} provided a statistical analysis on neural responses to object stimuli in primate anterior inferotemporal(AIT) cortex, and showed that the critical features for individual neurons in primate AIT cortex are not very complex, but there is an indefinitely large number of them, which is inconsistent with traditional structural models.
 Besides, based on the neural responses to object stimuli, they  found that the intrinsic dimensionality of object representations in AIT cortex is around 100  \citep{Lehky2014}.
Considering the current outstanding performance of DNNs in visual object recognition and their often claimed hierarchical structural emulation of primate ventral pathway, it is worthwhile investigating whether the responses of DNN neurons (units in DNNs) have similar response statistics to those of AIT neurons in primates. Following Lehky et al.'s works \citep{Lehky2011,Lehky2014},
in this work, we analyze the response statistics to image stimuli and the intrinsic dimensionality of the object representations of DNN neurons. Our findings show that
in terms of kurtosis and Pareto tail index, the response statistics on both single-neuron selectivity and population sparseness of DNN neurons are fundamentally different from those of IT neurons except some special cases. In addition, by increasing the number of neurons and image stimuli from
hundred-order to thousand-order, or even higher, the conclusions on neural responses could
alter substantially.
And with the ascendancy of the convolutional layers of DNNs, both the single-neuron selectivity and population
sparseness of DNN neurons increase, indicating that
the last convolutional layer is engaged in learning critical features for object representations,
while the following fully-connected layers are to learn prototypical categorization features. It is
also found that a sufficiently large number of stimulus images and sampled neurons, much larger than several hundreds in \citep{Lehky2014}, are necessary for obtaining a stable intrinsic dimensionality of object representations.
To the best of our knowledge, this is the first work to analyze the response statistics of DNN neurons comparing with primate IT neurons, and our results provide not only some insights into the discrepancy of DNN neurons with respect to primate IT neurons in object representation, but also shed some light on the possible outcomes of IT neurons when the  number of recorded neurons and stimuli is beyond current level of several hundreds
in \citep{Lehky2011,Lehky2014}.



\section{Introduction}

In recent years, deep neural networks (DNNs) have made tremendous progresses in many challenging tasks in the field of computer vision \citep{Simonyan2014,LeCun2015,Hinton2006,Szegedy2015,He2015,He2016,Zhang2016,Szegedy2015,Krizhevsky2012,Taigman2014}, especially in visual object recognition/classification.
DNNs have achieved comparable performances on some special recognition tasks with humans \citep{Simonyan2014,Szegedy2015,He2015,He2016}. One of the presumptions of their exceptionally good performances is the adopted hierarchical processing architecture, or a ventral-pathway-like visual processing structure in primates. This presumption is partially supported by Cadieu et al.'s work \citep{Cadieu2014}, where it was reported that DNNs could rival the representation of primate AIT neurons for core visual object recognition.
However, there is still a lack of insights to the differences and similarities of the object recognition mechanisms between DNNs and primate IT cortex at the level of neural responses.



On the neurophysiological side, it is generally believed that primate inferotemporal cortex is
the final stage in object recognition \citep{Gross2008}. Considering that activation of inferotemporal cells requires complex stimulus features \citep{Kobatake1994}, Lehky et al. \citep{Lehky2011} gave a statistical analysis on visual responses of monkey AIT neurons to object stimuli. In their work, the recorded neural responses were measured in two aspects: (i) single-neuron selectivity by measuring the responses of a single neuron to all the stimulus images; (ii) population sparseness by measuring the responses of all the neurons
stimulated by a single image. In addition, they also computed the intrinsic dimensionality of object representations in primate AIT cortex \citep{Lehky2014}, and showed that
it  is less than $100$.

In \citep{Lehky2011, Lehky2014}, the number of the used stimulus images is $806$ and the number of the sampled AIT neurons is $674$, which are both larger than those in many previous neurophysiological works \citep{Franco2007,Sirovich2009}. However, from the point of view of population response statistics, $806$ stimulus images and $674$ neurons seem not large enough to fully characterize the selectivity and sparseness of neural responses, as a monkey can recognize thousands of different objects, and the number of its IT neurons is around millions. It could be possible that the obtained response statistics
alter substantially when both the number of stimuli and the number of measured neurons are largely increased. The authors \citep{Lehky2014} were also aware of this issue, and pointed out that the major limitation of their analysis was the requirement to extrapolate far beyond the available data,
considering the infeasibility to measure too many neurons under too many stimuli.

Here, we would investigate the statistics on the responses of neurons in DNNs to visual object recognition, our goal are principally 3-fold: First, considering the widely claimed credit of DNNs' success to the ventral pathway-like layered structure, and the common belief that IT neurons in primates are primarily for representing objects, we would clarify the similarities and differences of the response statistics of neurons between DNNs and primate IT cortex; Second, since the number of the DNN neurons could be sufficiently large, our results on DNNs could act as an indicator to show the possible trends of
response statistics of AIT neurons if their number is much larger than the current level of hundreds; Third, we hope our results could shed some light on a possible reason of why DNNs excel on visual object recognition. In particular, the following issues are addressed:
\begin{itemize}
\item Using the same set of object stimuli as used in \citep{Lehky2011,Lehky2014}, are the statistical characteristics of neural responses in DNNs consistent with those in primate AIT cortex?
\item What are the response statistics of DNN neurons at different layers under different sizes of
stimulus image sets?
\item Does there exist an intrinsic dimensionality of object representation at the later layers of DNNs? If so, is it close to the intrinsic dimensionality of object representation in primate AIT cortex?
\end{itemize}

To answer these questions, we employ a popular deep neural network VGG  \citep{Simonyan2014} for our  experiments. The VGG network has achieved close performances to humans in the ILSVRC-2014 competition, and  has been extensively used as a feature extraction module in many existing works for tackling different vision tasks \citep{long2015fully,girshick2015fast}. The high layers in VGG are generally considered to provide object representations of input images. Specifically, we use both the set of object stimuli in \citep{ Lehky2011,Lehky2014} and the validation set (50000 images) from the ImageNet challenge dataset as our inputs to the VGG network,
to analyze the single-neuron selectivity and population sparseness of DNN neurons
at different layers by computing the kurtosis and Pareto tail index of neural responses. Then, the same methodology as used in \citep{Lehky2014} is adopted for estimating the intrinsic dimensionality of the output representations of DNN neurons.
 Finally, we provide a comparative analysis on the similarities and differences of response statistics between VGG and primate IT cortex.


\begin{figure}[t]
\centering
  \subfigure[Example images from \citep{Lehky2011,Lehky2014}] {  \includegraphics[height=6 cm]{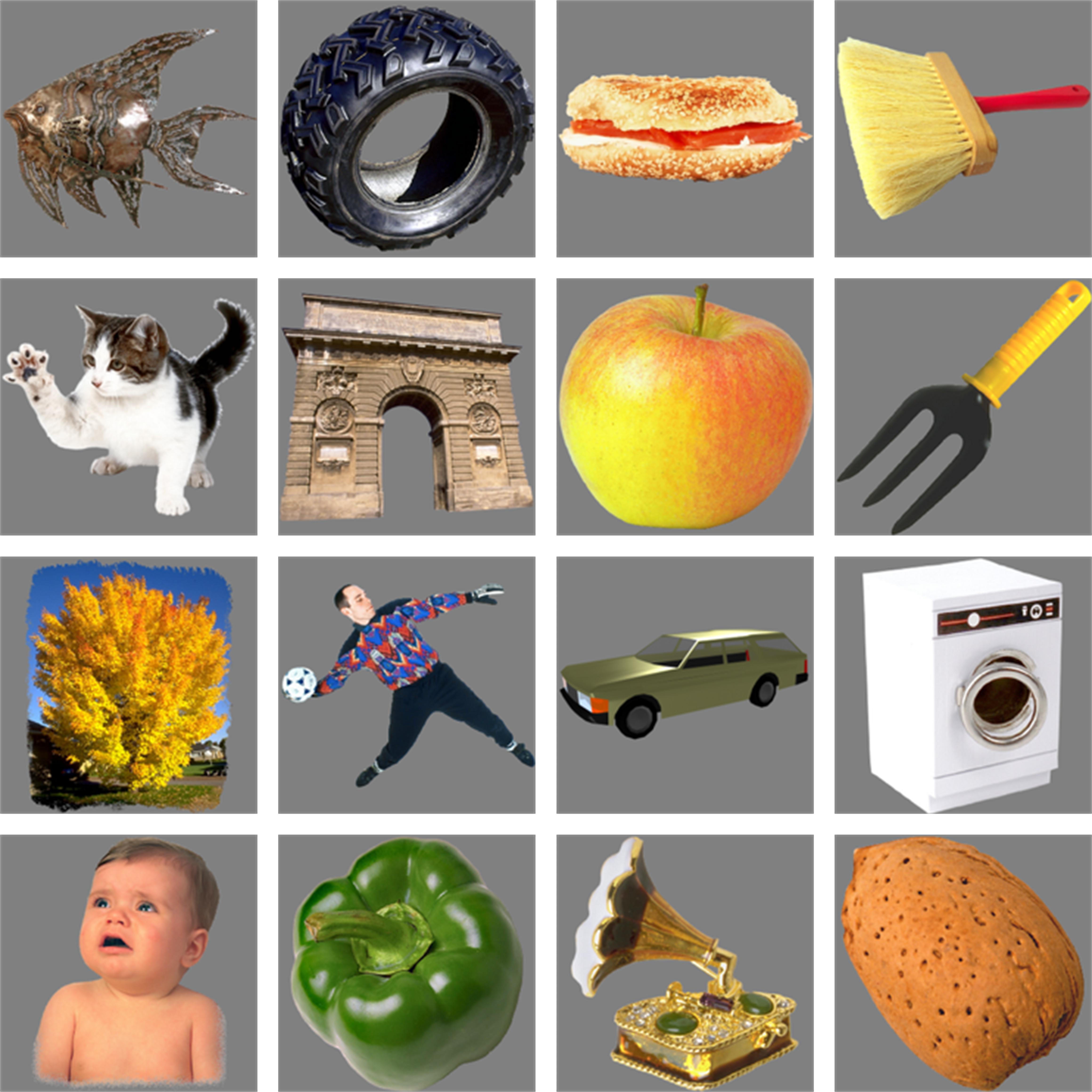}    \label{stimuli}   }
  \subfigure[Example images from the ImageNet dataset] {  \includegraphics[height=6 cm]{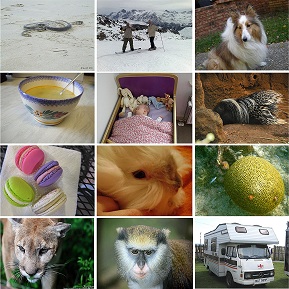}    \label{stimuli2}   }
\caption{Example stimulus images.}
\end{figure}

\section{Statistics of Visual Responses in Primate IT Neurons and Evaluation Criteria} \label{SIT}

Our goal is to use the same concepts and criteria to evaluate both the response statistics and the intrinsic dimensionality of object representations of DNN neurons,  as done in \citep{Lehky2011,Lehky2014} for the AIT neurons in monkey, here we at first introduce these concepts and criteria.

\subsection{Dataset}

In \citep{Lehky2011,Lehky2014}, the responses of $674$ AIT neurons of two macaque monkeys were recorded under $806$ stimuli, which were color images of natural and artificial objects.  Then, a $806\times 674$ neural response matrix was obtained. Each column was the responses of a single neuron to all the $806$ images, and each row was the responses of all the $674$ neurons to a single image.
 The size of each stimulus image is $125\times 125$, and several stimulus images are shown in Fig. \ref{stimuli}.

\subsection{Two Kinds of Response Probability Density Function}

Neural responses to a set of stimulus images, stored in the above mentioned neural response matrix, are statistically measured in two ways: One  is to measure single-neuron responses, which are the responses of a single neuron to all the stimulus images, i.e. the column vectors in the neural response matrix; The other one is to measure
population responses, which are the responses of all the neurons stimulated by a single stimulus image, i.e. the row vectors of the neural response matrix.

\begin{figure}[t]
\centering
  \subfigure {  \includegraphics[height=4 cm]{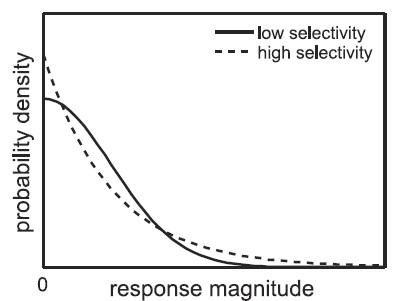}  }
\caption{Probability density functions for high-selectivity and low-selectivity responses \citep{Lehky2011}.} \label{myPDFs}
\end{figure}

Accordingly, each column of the neural response matrix is fitted into a single-neuron response probability density function, which is used later to evaluate the single-neuron selectivity. Similarly, each row is also fitted into a population response probability density function to evaluate the population sparseness.
Fig. \ref{myPDFs}  (from \citep{Lehky2011}) shows an example of selectivity probability density functions, where the high-selectivity probability density function (dashed line) has a heavier upper tail than the low-selectivity probability density function (solid line).

\subsection{Single-Neuron Selectivity, Population Sparseness, and Two Evaluation Criteria} \label{SSTC}

Single-neuron selectivity and population sparseness are two related characteristics to reflect the efficiency and capacity of object representations in AIT cortex,
which are extensively studied in many neurophysiological works \citep{Franco2007,Lehky2005,Lehky2011}.
The single-neuron selectivity reflects the response distributions of individual neurons to
different stimuli,
while the population sparseness reflects  the response distributions of all the neurons to a single image.

For the above mentioned $806\times 674$ neural response matrix, each neuron has a fitted selectivity response probability density function, and totally $674$ such functions for all the
$674$ neurons are used to evaluate the single-neuron selectivity. Similarly,
each stimulus image has a population response probability density function, and totally $806$ such functions for all the $806$ stimulus images are used to evaluate the population sparseness.

Given a probability density function for single-neuron responses (or population responses) as shown in Fig. \ref{myPDFs}, if it has a substantial upper tail, it means high-selectivity responses have a
larger probability of occurrence at the extreme, in other words, the neuron is more selective (or the population response is more sparse).   Hence, the single-neuron selectivity (population sparseness) can be measured by ``upper-tail-heaviness"  of the probability density function.
In \citep{Lehky2011}, this upper-tail heaviness was evaluated by two criteria, one is the kurtosis, and the other is the Pareto tail index.
The kurtosis and the Pareto tail index are discussed in Appendix A.

\subsection{Intrinsic Dimensionality of Object Representations} \label{IDC}

Object representations in AIT neurons are largely redundant, it is speculated that some
``intrinsic dimensionality" exists for such representations \citep{Lehky2014}.
Loosely speaking, among such redundant representations, some ``independent" representations
should exist.

In  \citep{Lehky2014},  Lehky et al. used both the Grassberger-Procaccia algorithm \citep{Grassberger1983} and a PCA(Principal Component Analysis)-based method to estimate the intrinsic dimensionality of object representations, and found the PCA-based method is more
stable. Hence, in this work, we  use only the PCA-based method to compute
the intrinsic dimensionality of the object representations of DNN neurons.

Broadly speaking, the intrinsic dimensionality computation by PCA is done in a 2-step
manner. Firstly, the intrinsic dimensionality of neuron responses under a fixed number
of neurons and image stimuli is computed by comparing the eigenvalue rankings of the
original response matrix to their reshuffled one, then
the final intrinsic dimensionality is computed as the asymptotic one by extrapolation under
different combinations of image number and neuron number.
The detailed  PCA-based method for the intrinsic dimensionality estimation is provided in Appendix B.

%
%
%



\begin{figure}[t]
\centering
  \subfigure{  \includegraphics[height=2.3 cm]{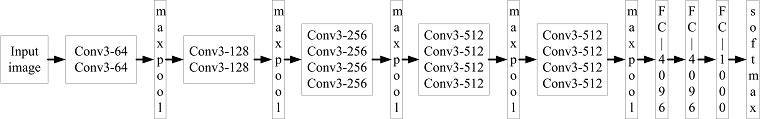}    \label{VGG}   }
\caption{Architecture of VGG where the
convolutional layer parameters are denoted as ``$\langle$Conv receptive field size$\rangle$ - $\langle$number of channels$\rangle$''.
}
 \label{VGG}
\end{figure}

\section{Response Statistics of DNN Neurons and Intrinsic Dimensionality}
\label{sec:1}

The VGG network \citep{Simonyan2014}, a widely used deep neural network, is employed as our model
DNN. Then, the methods
and criteria in Section \ref{SIT} are used to evaluate the
response statistics of DNN neurons at different layers and the corresponding intrinsic dimensionality of object representations.

\subsection{Datasets and VGG}

We use the following two datasets as object stimuli in our experiments.
The first one is the dataset used in \citep{Lehky2011,Lehky2014}, including $806$ images as described in Section
\ref{SIT}, denoted as Dataset I.
Since a small set of samples may miss rare large events, then underestimate
single-neuron selectivity and population sparseness of neural responses,
a larger dataset -- the validation set from the ImageNet dataset -- is also utilized, denoted as Dataset II.
Dataset II includes $50000$ images belonging to 1000 classes, 50 images per class.  Several example images
from this dataset are shown in Fig. \ref{stimuli2}.
All the images in the two datasets are resized to $224\times 224$ RGB images.

The VGG network, consisting of $16$ convolutional layers with $3\times 3$ convolution filters
and $3$ fully-connected layers, was originally designed for object categorization.
Its input is
a fixed-size $224\times 224$ RGB image, and its architecture is shown in Fig. \ref{VGG}.
Considering that it is still pendent which layer in a deep network acts similarly as primate IT cortex does,
the statistics of the neural responses at the last seven layers (including four convolutional layers and three fully-connected layers) in VGG are computed.
Note that the last layer of VGG outputs a probability vector of object category, and the other layers output different features.

 For notational convenience, the last seven layers are denoted as $\{L1, L2, L3, L4, L5, L6, L7\}$ respectively. As noted from Fig. \ref{VGG}, the $\{L1, L2, L3, L4\}$ layers are the convolutional layers, each of which has $100352=14\times 14\times 512$ neurons. The $\{L5, L6, L7\}$ layers are the fully-connected layers. The $\{L5, L6\}$ layers have $4096$ neurons respectively, and the $L7$ layer has $1000$ neurons.

\begin{figure}[!t]
\centering
\subfigure[] {  \includegraphics[height= 7.2 cm]{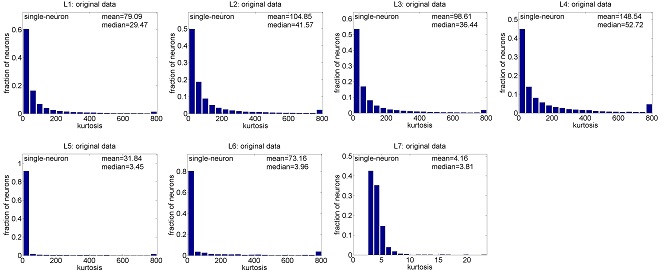} \label{fseleca}}
\subfigure[] {  \includegraphics[height= 7.2 cm]{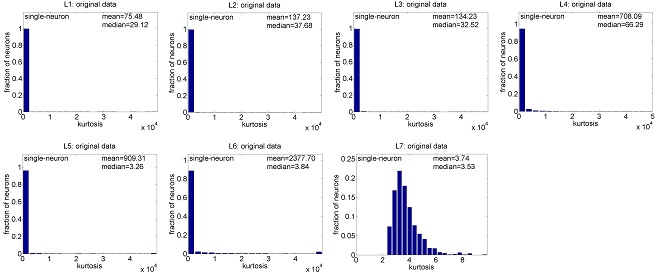} \label{fselecb}}
\caption{Single-neuron selectivity of neural responses at different layers of VGG to Dataset I and Dataset II:
(a) Selectivity for Dataset I; (b) Selectivity for Dataset II.} \label{fselecD1}
\end{figure}

\subsection{Neural Response Statistics by Kurtosis} \label{CSS}

\paragraph{3.2.1 Single-Neuron Selectivity \\}

For each neuron at the last seven layers of VGG, the kurtosis of its responses to the $806$ images in Dataset I and the kurtosis of its responses to the $50000$ images in Dataset II are computed respectively for measuring the single-neuron selectivity, and the corresponding results are shown in Fig. \ref{fselecD1}. Since normalization does not affect the single-neuron selectivity as discussed in Appendix A, we do not show the computed kurtosis of the normalized responses.

As seen from Fig. \ref{fselecD1},  for
the layers  $\{L1,L2,L3,L4,L5,L6\}$, there exist
a few neurons whose kurtosis values are of one or two orders larger
than median kurtosis value, indicating that these
neurons are much sensitive to a few special stimuli from the two datasets.

For the four convolutional layers $\{L1,L2,L3,L4\}$ which have the same number of neurons, both the computed mean and median kurtosis on Dataset I and Dataset II tend to become larger  with the increase of the layer number (although the mean and median kurtosis at Layer $L2$ are slightly larger than those at Layer $L3$).   This demonstrates that
for different convolutional layers with the same number of neurons, the probability distribution of the single-neuron responses of the neurons at a higher layer may have a heavier upper tail than that at a lower layer.


For the fully-connected layers $\{L5,L6\}$, the computed mean and median kurtosis values on the two datasets at Layer $L5$ are slightly smaller than those at Layer $L6$.
Moreover, the computed mean and median kurtosis values at the two fully-connected layers $\{L5,L6\}$ are lower than or close to those
at the last convolutional layer $L4$ on the two datasets in most cases. This demonstrates that the probability distribution of the single-neuron responses at
the two fully-connected layers $\{L5,L6\}$ has a lighter or close upper tail than that at the last convolutional layer.

For the last fully-connected layer $L7$,  the computed mean and median kurtosis values on Dataset II are close to those on Dataset I, and they are lower than those at the other layers, which means most of the neurons at this layer have weak selectivity.
This is quite an expected result, because this layer outputs a probability vector of object category, and it is not for object representation.

\paragraph{3.2.2 Population Sparseness \\}  \label{CSS2}

\begin{figure}[!t]
\centering
\subfigure[] {  \includegraphics[height= 7.2 cm]{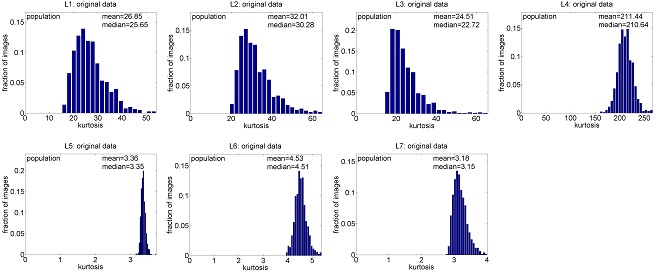} \label{fpopuD1a}}
\subfigure[] {  \includegraphics[height= 7.2 cm]{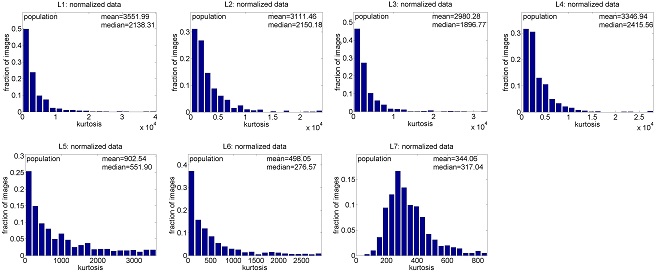} \label{fpopuD1b}}
\caption{Population sparseness of responses at different layers of VGG to Dataset I:
(a)
Sparseness for the unnormalized data; (b) Sparseness for the normalized data.} \label{fpopuD1}
\end{figure}

\begin{figure}[!t]
\centering
\subfigure[] {  \includegraphics[height= 7.2 cm]{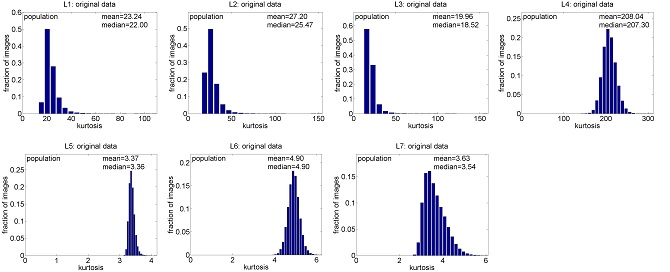}  \label{fpopuD2a} }
\subfigure[] {  \includegraphics[height= 7.2 cm]{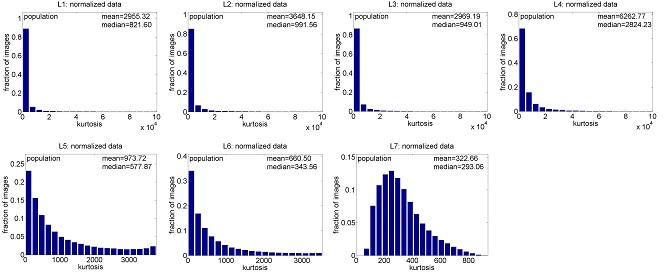}  \label{fpopuD2b} }
\caption{Population sparseness at different layers of VGG to Dataset II: (a) Sparseness for the unnormalized data;
(b) Sparseness for the normalized data.} \label{fpopuD2}
\end{figure}

For each image in Dataset I and Dataset II, the kurtosis of the population responses at each layer is computed for measuring the population sparseness, and the results are shown in Fig. \ref{fpopuD1a} and Fig. \ref{fpopuD2a}. In addition to the raw data,
we also perform the sparseness calculation with the normalized data via Eq. (\ref{Knorm}) in Appendix A, and the kurtosis values for the  normalized population responses are reported in Fig. \ref{fpopuD1b} and Fig. \ref{fpopuD2b}.

\paragraph{Sparseness for the Unnormalized Data \\}

As seen from Fig. \ref{fpopuD1a} and Fig. \ref{fpopuD2a},
for the four convolutional layers $\{L1,L2,L3,L4\}$, the computed mean and median kurtosis values on Dataset I and Dataset II tend to become larger with the increase of the layer number (although the mean and median kurtosis at Layer $L2$ are slightly larger than those at Layer $L3$), which is  consistent with the hierarchical feature binding nature.   This demonstrates that
for different convolutional layers with the same number of neurons, the probability distribution of the population responses of the neurons at a higher layer has a heavier upper tail than that at a lower layer.


For the fully-connected layers $\{L5,L6\}$, the computed mean and median kurtosis values at Layer $L5$ are slightly smaller than those at Layer $L6$.
But, the computed mean and median kurtosis values at Layers $\{L5,L6\}$ are much lower than those
at the last convolutional layer $L4$ on the two datasets.
The computed mean and median kurtosis values  at Layer $L7$ are lower than those at the other layers in most cases.

The above results seem to suggest that the DNN neurons at Layer $L4$ are responsible
for object representation. From Layer $L4$ to Layer $L7$, the
object representations in the same category gradually lost their individualism,
and the prototypical categorization features are formed in a layer-by-layer fashion.

\paragraph{Sparseness for the Normalized Data \\}

As seen from Fig. \ref{fpopuD1a}, Fig. \ref{fpopuD1b}, Fig. \ref{fpopuD2a}, and Fig. \ref{fpopuD2b}, the mean and median kurtosis values for the normalized population responses are much larger than those for the unnormalized population responses, mainly because some neural responses are amplified unusually when they are normalized with different small mean responses respectively,
due to the fact that for some images, many neurons are not responsive, or of very weak
response.

For the four convolutional layers $\{L1,L2,L3,L4\}$, the changing trend of the computed mean and median kurtosis for the normalized data is similar to that for the unnormalized data,
confirming once again that
for different convolutional layers with the same number of neurons, the population sparsity at a higher layer is generally stronger than that at a lower layer.

For the fully-connected layers $\{L5,L6\}$, the computed mean and median kurtosis values for the
normalized data at Layer $L5$ are larger than those at Layer $L6$, but the computed mean and median kurtosis values at these two layers are much lower than those
at the last convolutional layer $L4$ on both the two datasets.
The computed mean and median kurtosis values for the normalized population responses at Layer $L7$ are lower than those at the other layers in most cases.
The above results are also consistent with the categorization nature
from Layer $L4$ to Layer $L7$ from the viewpoint of the population responses.

\paragraph{3.2.3 Variations of Response Statistics by Changing Neuron Numbers and Stimulus Numbers \\} \label{EDSSS}

\begin{figure}[!t]
\centering
\subfigure[] {  \includegraphics[height= 7.2 cm]{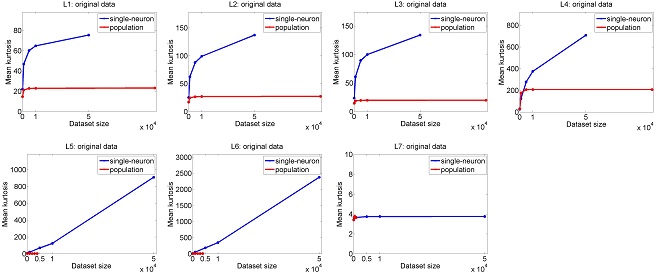} \label{feffectD2a}}
\subfigure[] {  \includegraphics[height= 7.2 cm]{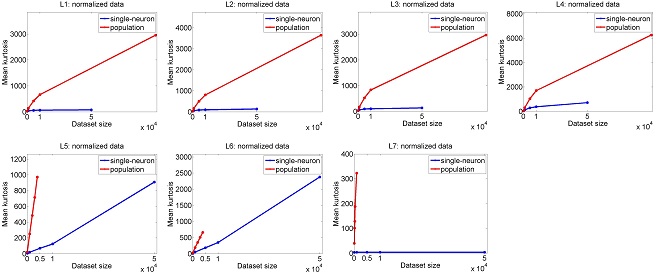} \label{feffectD2b}}
\caption{Mean kurtosis for both single-neuron responses and population responses with different image samples and neurons at different layers of VGG: (a) For the unnormalized data; (b) For the normalized data.} \label{feffectD2}
\end{figure}

\begin{figure}[!t]
\centering
\subfigure[] {  \includegraphics[height= 7.2 cm]{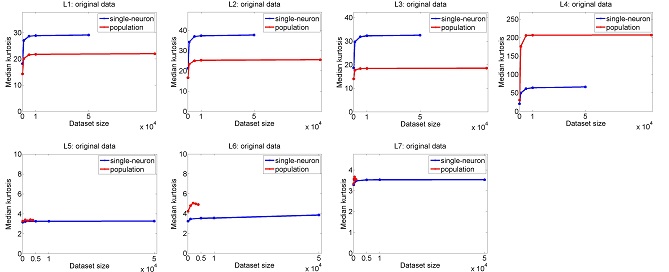} \label{feffectD2Ma}}
\subfigure[] {  \includegraphics[height= 7.2 cm]{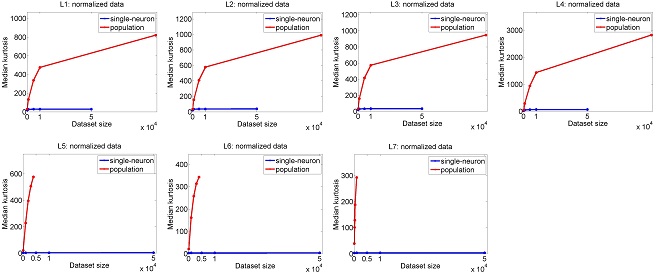} \label{feffectD2Mb}}
\caption{Median kurtosis for both single-neuron responses and population responses with different image samples and neurons at different layers of VGG: (a) For the unnormalized data; (b) For the normalized data.} \label{feffectD2M}
\end{figure}

Here, we calculate the single-neuron selectivity and the population sparseness of  responses as the dataset size increases by resampling subsets from Dataset II. Note that as done in \citep{Lehky2011}, when dealing with the single-neuron responses, dataset size refers to the number of the stimulus images tested on each neuron. When dealing with the population responses, dataset size refers to the number of the DNN neurons in the population.

The image subset size of $[100, 1000, 5000, 10000, 50000]$ is evaluated respectively. The neuron subset sizes for each of the layers $\{L1,L2,L3,L4\}$ are $[100, 1000, 5000, 10000, 100000]$, the neuron subset sizes for both the layers $\{L5,L6\}$ are $[100, 1000, 2000, 3000, 4000]$, and the neuron subset sizes for Layer $L7$ are $[100, 200, 300, 400, 1000]$. Under each combination of the image number and the neuron number, the sampling is independently done 10 times, and the mean value of the 10 estimated kurtosis values is used as the final kurtosis.
Fig. \ref{feffectD2} shows the mean kurtosis values with different image samples and neurons at different layers of VGG, and Fig. \ref{feffectD2M} shows the median kurtosis values with different image samples and neurons at different layers of VGG.

As is seen, both
single-neuron selectivity and population sparseness behave in a monotonically non-decreasing order  as the size of the dataset increases. For the original unnormalized data, population sparseness increases slower than single-neuron selectivity, but for the normalized data, population sparseness increases faster than single-neuron selectivity.
For the subsets of different sizes, the changing trends of their computed mean kurtosis from Layer $L1$ to Layer $L7$ are similar, and the changing trends of their computed median kurtosis from Layer $L1$ to Layer $L7$ are also similar.

\subsection{Neural Response Statistics by Pareto Tail Index}

\begin{figure}[!t]
\centering
\subfigure[] {  \includegraphics[height=7.2 cm]{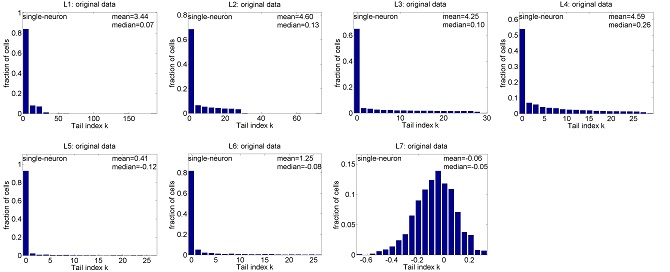} }
\subfigure[] {  \includegraphics[height=7.2 cm]{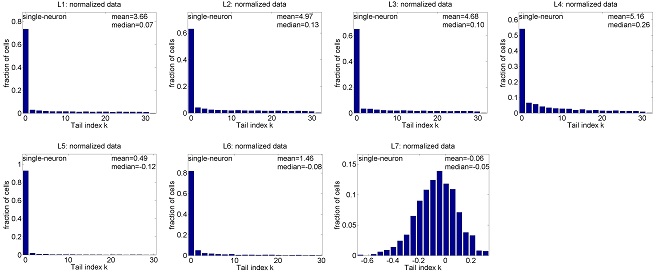} }
\caption{Histograms of the computed Pareto tail indices of the single-neuron responses to Dataset I:(a) For the unnormalized data; (b) For the normalized data.} \label{ftailselecD1}
\end{figure}

\begin{figure}[!t]
\centering
\subfigure[] {  \includegraphics[height= 7.2 cm]{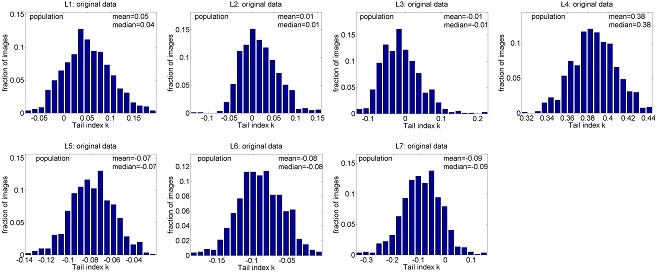} }
\subfigure[] {  \includegraphics[height= 7.2 cm]{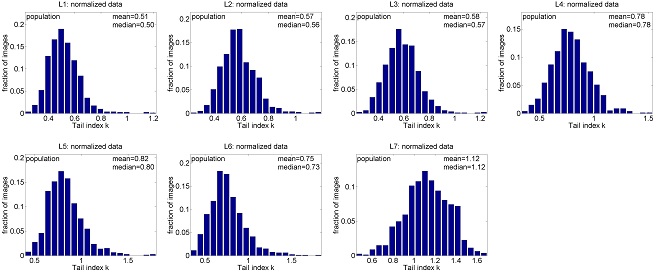} }
\caption{Histograms of the computed Pareto tail indices of the population responses to Dataset I:(a) For the unnormalized data; (b) For the normalized data.} \label{ftailpopuD1}
\end{figure}

\begin{figure}[!t]
\centering
\subfigure[] {  \includegraphics[height= 7.2 cm]{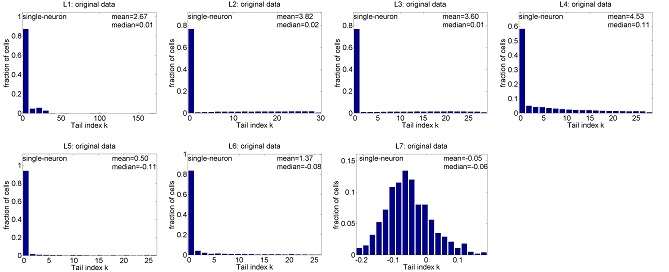} }
\subfigure[] {  \includegraphics[height= 7.2 cm]{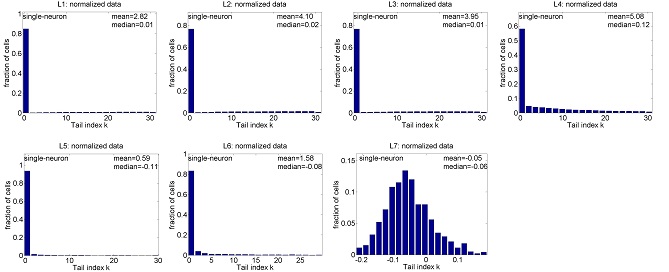} }
\caption{Histograms of the computed Pareto tail indices for the single-neuron responses to Dataset II:(a) For the unnormalized data; (b) For the normalized data.} \label{ftailselecD2}
\end{figure}

\begin{figure}[!t]
\centering
\subfigure[] {  \includegraphics[height= 7.2 cm]{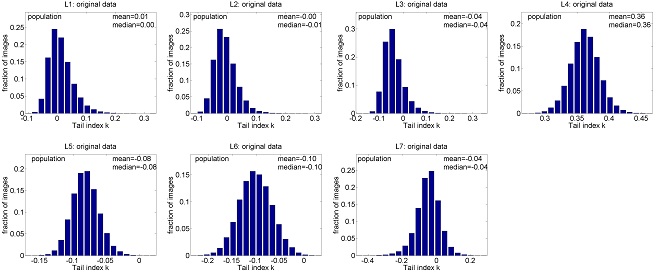} }
\subfigure[] {  \includegraphics[height= 7.2 cm]{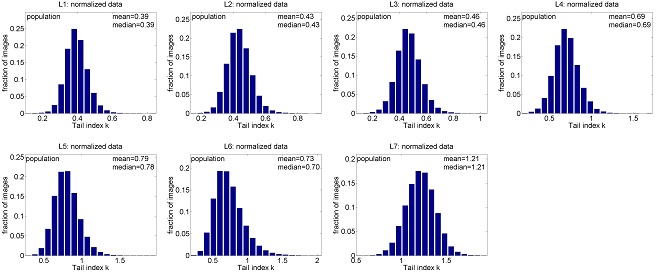} }
\caption{Histograms of the computed Pareto tail indices for the population responses to Dataset II:(a) For the unnormalized data; (b) For the normalized data.} \label{ftailpopuD2}
\end{figure}


At each of the last seven layers of DNNs, generalized Pareto distributions (GPDs) are fitted for both the probability distribution function of single-neuron responses and the probability distribution function of the population responses. Fig. \ref{ftailselecD1} and Fig. \ref{ftailpopuD1} show the histograms of the computed Pareto tail index(i.e. $k$ in Eq. (\ref{Pareto}) of Appendix A) for the
single-neuron responses and the population responses to the stimulus images from Dataset I.
Fig. \ref{ftailselecD2} and Fig. \ref{ftailpopuD2} show the histograms of the computed Pareto tail index for the single-neuron responses and the population responses to the stimulus images from Dataset II.

\paragraph{3.3.1 Tail Index for Single-Neuron Responses \\}

As seen from Fig. \ref{ftailselecD1} and Fig. \ref{ftailselecD2}, for all the referred layers, the computed mean and median values of $k$ for the normalized single-neuron responses is quite close to those for the unnormalized single-neuron
responses.

For the four convolutional layers $\{L1,L2,L3,L4\}$, with the increase of the layer number, the mean and median values of $k$ on the two datasets tend to increase (although the mean and median values of $k$ for both the normalized/unnormalized data at Layer $L2$ is slightly larger than those at Layer $L3$), which is consistent with the corresponding kurtosis results on the single-neuron responses in Section \ref{CSS}.
 For the fully-connected layers $\{L5,L6\}$, the mean and median values of $k$  for both the normalized/unnormalized data at Layer $L5$ are slightly smaller than those at Layer $L6$.
And the computed mean and median values of $k$ at $\{L5,L6\}$ are lower than those
at the last convolutional layer $L4$.

\paragraph{3.3.2 Tail Index for Population Responses \\}

As seen from Fig. \ref{ftailpopuD1}  and Fig. \ref{ftailpopuD2}, for the four convolutional layers $\{L1,L2,L3,L4\}$, with the increase of the layer number, the mean and median values of $k$ for the population responses tend to become larger, although the mean and median values of $k$ at Layer $L2$ is slightly larger than those at Layer $L3$, which is consistent with the above kurtosis results on the population responses.

For the fully-connected layers $\{L5,L6\}$, the mean and median values of $k$ for the unnormalized data at Layer $L5$ are slightly larger than those at Layer $L6$, which is inconsistent with the corresponding kurtosis results. The mean and median values of $k$ for the normalized data at Layer $L5$ are also slightly larger than those at Layer $L6$, which is consistent with the corresponding kurtosis results.
The computed mean and median values of $k$ data at $\{L5,L6\}$ are also lower than or close to those at the last convolutional layer $L4$.
In addition, the mean and median values of $k$ at Layer $L7$ are close to or larger than those at the other layers, which is not consistent with the corresponding kurtosis results on the population responses.

Moreover, for each referred layer, the mean and median values of $k$ for the normalized population responses are larger than those for the unnormalized population responses,
which is consistent with the  corresponding kurtosis results.  The mean value of $k$ for the normalized/unnormalized population responses  to the two datasets is lower than that for the single-neuron responses at each of Layers $\{L1,L2,L3,L4,L5,L6\}$ in most cases.  
 The mean value of $k$ for the unnormalized population responses to Dataset I at Layer $L7$ is also lower than that for the single-neuron responses, while the mean value of $k$ for the normalized/unnormalized population responses to Dataset II at Layer $L7$ is slightly larger than that for the single-neuron responses.

\subsection{DNN Neurons Versus AIT Neurons on Statistics of Responses} \label{discussion}

In \citep{Lehky2011}, kurtosis and tail index were used respectively to measure single-neuron selectivity and
population sparseness of AIT neurons, the main results include:
(i) For the unnormalized neural responses, the  population sparseness measured by kurtosis is greater than
the single-neuron selectivity;
(ii) For the normalized neural responses, the population sparseness measured by kurtosis is also greater than
the single-neuron selectivity;
(iii) The mean tail index for the unnormalized population responses is greater than
that for the unnormalized single-neuron responses;
(iv) The mean tail index for normalized population responses is greater than
that for the normalized single-neuron responses;
(v) The observations (i)--(iv) demonstrate that the estimated probability distributions for the single-neuron responses in primate AIT cortex have lighter tailes, indicating that the critical features for individual neurons in primate AIT cortex are not quite complex. In contrast, the estimated probability distributions for the population responses have heavier tails, indicating  that there is an indefinitely large number of different critical features. These results are inconsistent
with the traditional structural model of object recognition where a small number of
standard features is used.

In comparison, three main points are revealed in our above experimental results:

(1)  It is observed that
(i) For the unnormalized responses of DNN neurons at each of the last seven layers,
the population sparseness measured by kurtosis is smaller than
the single-neuron selectivity in most cases;
(ii) For the normalized responses of DNN neurons at each of the last seven layers,
the population sparseness measured by kurtosis is much greater than
the single-neuron selectivity in most cases;
(iii) The mean tail index for the unnormalized population responses is smaller than
that for the unnormalized single-neuron responses in most cases;
(iv) The mean tail index for the normalized population responses is also smaller than
that for the normalized single-neuron responses in most cases.
Comparing these results with the results of AIT neurons in \citep{Lehky2011},
except for the population sparseness and the single-neuron selectivity measured by kurtosis for the normalized data,  the conclusions on the population sparseness versus the single-neuron selectivity for AIT neurons in \citep{Lehky2011} are in direct conflict to those of DNN neurons in this work, that is to say, the population sparseness is, in contrast,  smaller than the corresponding single-neuron selectivity for DNN neurons. This means the DNN neurons are quite selective, and the population responses of DNN neurons are not as sparse as those by AIT neurons, indicating that the object representation of DNN neurons is fundamentally different from that of AIT neurons in monkey. Considering the ubiquitous cross-talks between the ventral and dorsal pathways, substantial feedbacks from higher visual areas to lower ones, and omnipresence of horizontal inhibitions, the discrepancy of object representations between the AIT and DNN neurons is not surprising. However, such direct conflicting results revealed in this work are nevertheless unexpected.

For the kurtosis measure for the normalized responses of DNN neurons, the single-neuron selectivity is smaller than the population sparseness, which is consistent with the AIT neurons, but contrary to the results measured by Pareto tail index for both the normalized and unnormalized data as well as the results by kurtosis for the unnormalized data.
The possible reason of this discrepancy could be 2-fold: At first, due to the 4-degree computational nature of kurtosis as defined in (\ref{Kurtosis}) of Appendix A, it is rather sensitive to noise. We suspect this discrepancy is caused by noise; Secondly, the kurtosis is a global measure of the probability density function (PDF) of neuron responses, it is possible in theory for a given set of neural responses, the reached conclusion by kurtosis is different from that by Pareto tail index. Since the Pareto tail index is designed specifically for measuring the tail heaviness of PDFs, we thought the results by tail index are more trustworthy for comparison to the results in \citep{Lehky2011}.

(2)  Our results show that the statistics of the neural responses at the convolutional layers are largely different from those at the fully connected layers. When the convolutional layer number increases,
both the kurtosis and tail index measures of single-neuron selectivity and population sparseness for the unnormalized/normalized responses increase, indicating that with the increase of the convolutional layer number, the critical feature for individual neurons in DNNs would become more complex, and there would be more features tuned by different neurons. This appears a typical hierarchical feature binding process.
The kurtosis and tail index measures of single-neuron selectivity and population sparseness at the last convolutional layer are also larger than those at the fully-connected layers in most cases.
Therefore, the responses at the last convolutional layer can be considered as the object representation. From then on, the fully connected layers are mainly engaged in learning a relatively smaller set of prototypical features for object categorization, and the final one is the categorization output.
We thought a good object representation should at least satisfy the following two purposes: First for object identification which needs to discriminate subtle differences from each other; second for object categorization which removes the individualism within the same category, hence the significant differences between the convolutional layers and the fully-connected layers revealed in this work are insightful and reasonable. For example, in the HMAX model \citep{Riesenhuber1999}, the categorization and object recognition submodels share the same input, or the same output of object representation.

(3) With the increase of the image number, it is noted from Fig. \ref{feffectD2} and Fig. \ref{feffectD2M}: (i) The estimated kurtosis for the single-neuron responses at each referred layer increases smoothly when the image number is lower than $1000$, and these results are consistent with those in
AIT neurons \citep{Lehky2011} where $806$ images are used as the stimuli to AIT neurons. (ii) When the image number is large than $10000$, the estimated kurtosis for the single-neuron responses at the convolutional layers reaches a stable value, and those at the fully-connected layers(except Layer $L7$) increase fast. These results are not consistent with those in
AIT neurons \citep{Lehky2011}, and they might indicate that the statistics of single-neuron responses in IT cortex based on only $806$ images cannot adequately characterize their single-neuron selectivity. To reach a stable result, more images (in our case, larger than $10000$ at least) are necessary.

When the neuron number increases, it is found from Fig. \ref{feffectD2} and Fig. \ref{feffectD2M}: (i) The estimated kurtosis for the unnormalized population responses at each referred layer increases smoothly, and approximately reaches a stable value. The results with the neuron number lower than $1000$ are consistent with those in
AIT neurons \citep{Lehky2011} where the responses of $674$ neurons are recorded;
(ii) The estimated kurtosis for the normalized population responses at each referred layer increases, and their increase decelerates with the increase of the neuron number. These results are inconsistent with those in AIT neurons where the increase of the mean kurtosis for the normalized population responses of AIT neurons is accelerated when the neuron number increases from $4$ to $650$. This might indicate that the statistics of neural responses in IT cortex based on only less than $1000$ neurons cannot adequately characterize their population sparseness. To reach a stable result, more neurons (in our case about $10000$) are necessary.

\subsection{Intrinsic Dimensionality} \label{CID}

\paragraph{3.5.1 Calculating Intrinsic Dimensionality by the PCA-based Method \\}

According to the discussions in Section \ref{discussion},
the features learnt from the last convolutional layer $L4$ of VGG are more appropriate for object
representations, and the features learnt from the first two fully-connected layers
are used for object categorization.
Hence in this section, the intrinsic dimensionality of the outputted object features at the layers $\{L4,L5, L6\}$  of VGG are  computed by the PCA-based estimation method, in order to
further investigate the characteristics of object representation of neural responses in DNNs.

\begin{figure}[t]
\centering
\subfigure[] {  \includegraphics[height=3.4 cm]{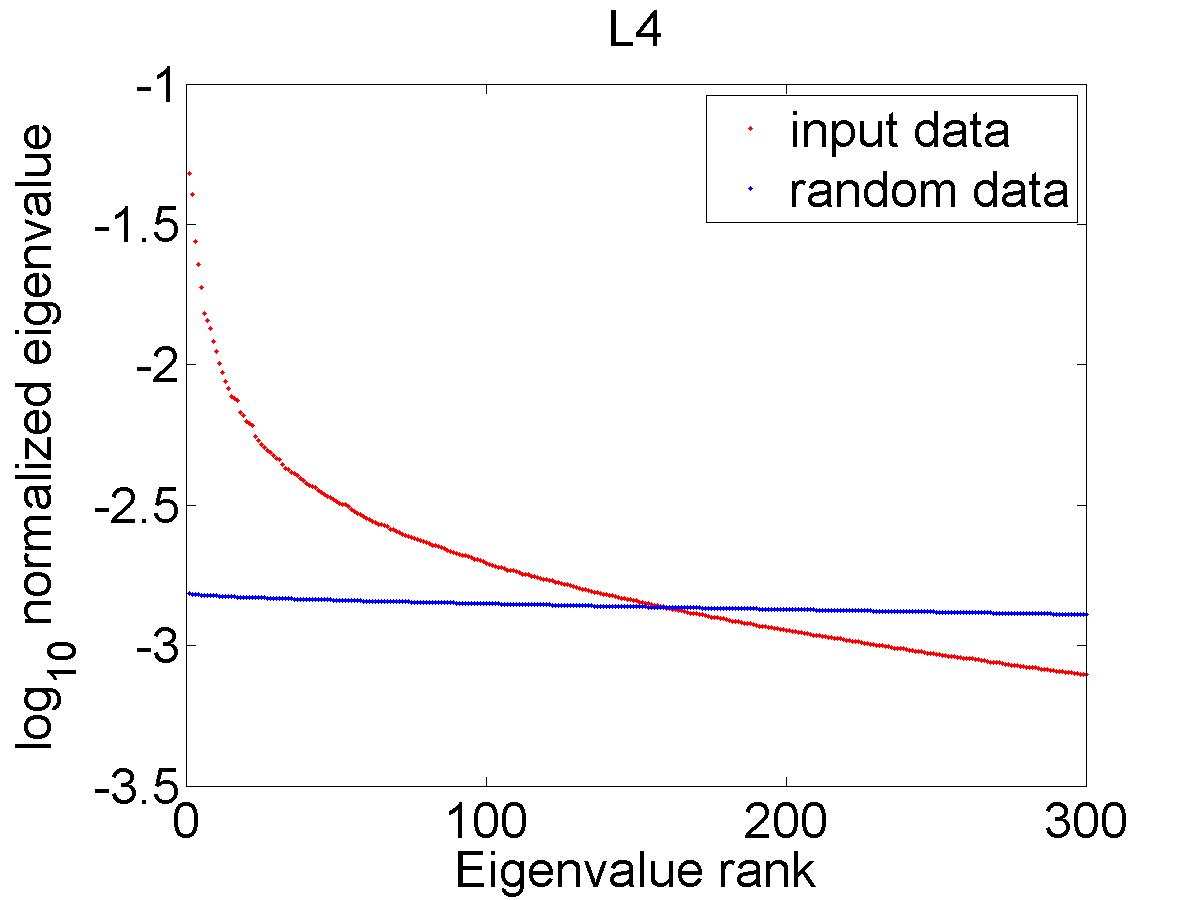} }
\subfigure[] {  \includegraphics[height=3.4 cm]{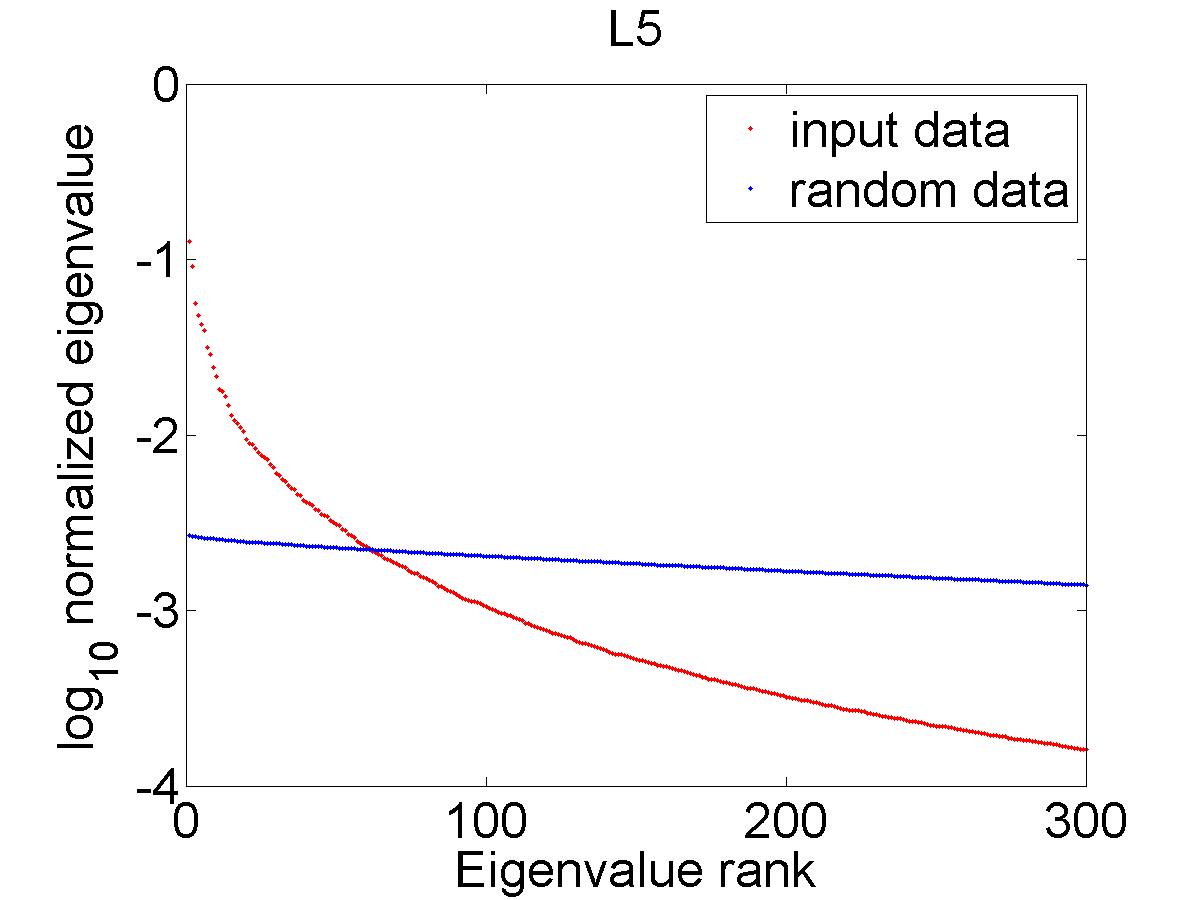} }
\subfigure[] {  \includegraphics[height=3.4 cm]{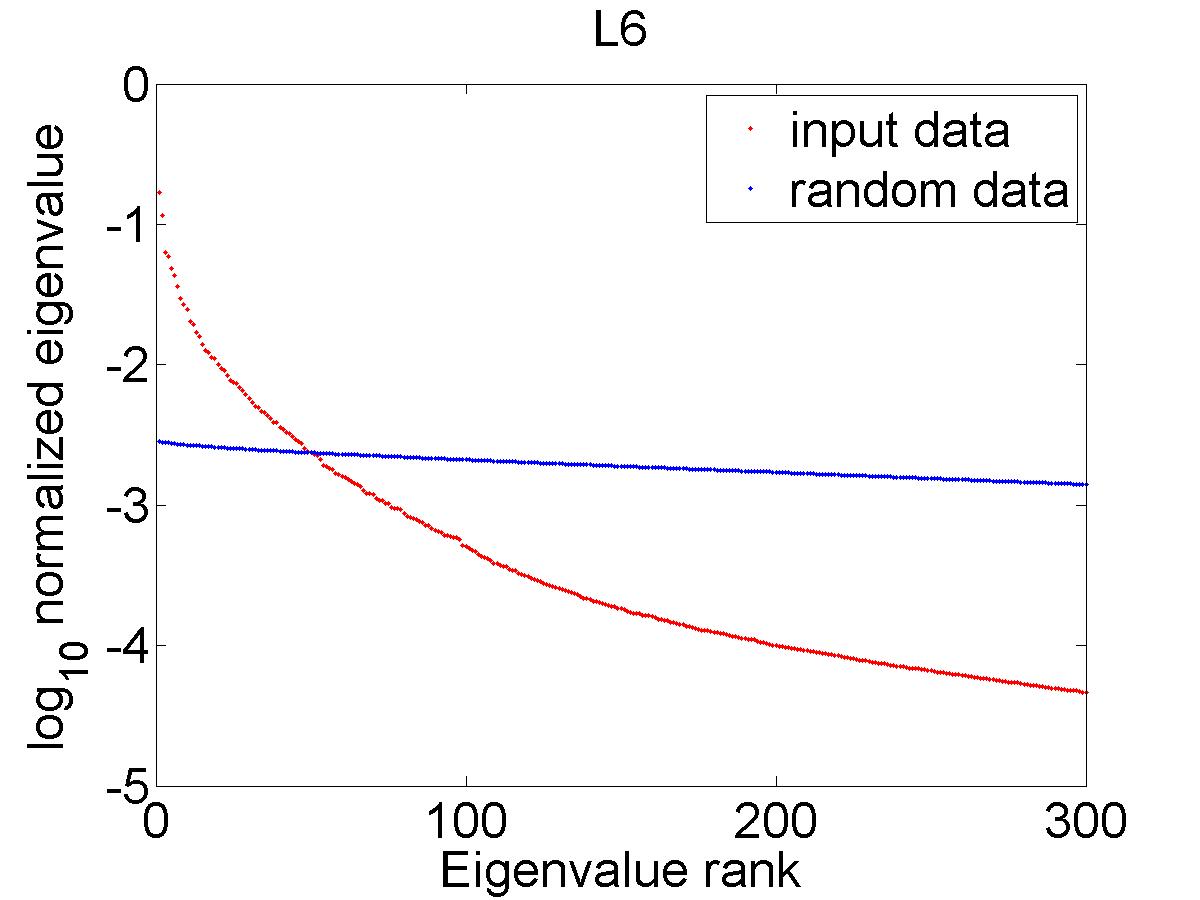} }
\caption{Examples on estimating intrinsic dimensionality with the responses at Layers $\{L4,L5,L6\}$ for Dataset I.}\label{IntrinsicDim}
\end{figure}

Fig. \ref{IntrinsicDim} shows three example results on estimating the intrinsic dimensionality of the object representation of neurons at the three layers with the stimulus images from Dataset I, where the red curve represents the rank-ordered eigenvalues of the original responses,  and the blue curve represents the rank-ordered eigenvalues of randomly reshuffled responses.
The intersection point of the two curves in each subfigure is the estimated dimensionality $\{159,62,50\}$  for Layers $\{L4, L5, L6\}$ respectively. In addition, the first two principal components of the corresponding response matrix for Layer $L4$ accounts for $9\%$ of the variance, lower than $17\%$ for the inferotemporal responses to Dataset I \citep{Lehky2014}, and $15\%$ for the inferotemporal responses obtained in \citep{Baldassi2013}, probably because
the number of neurons at Layer $L4$ is much larger than those in \citep{Lehky2014}
and \citep{Baldassi2013}.
The first two principal components of the corresponding response matrices for Layers $\{L5, L6\}$ respectively account for $\{22\%, 29\%\}$ of the variance. This suggests that with the increase of
the fully-connected layer number, the learnt features  become more prototypical for categorization.

Theoretically, the intrinsic dimensionality should be independent of the dataset size as well as the
number of the recorded neurons. However, in practice, any estimation depends on the used dataset size and the number of the recorded neurons. To address this problem, as done in \citep{Lehky2014},  we also compute the intrinsic dimensionality over a two-dimensional grid using neuron subsets of different sizes and image subsets of different sizes. The dimensionality, as a function of the image number and the neuron number, can be plotted as a surface in a three-dimensional space.

\begin{figure}[!t]
\centering
\subfigure[] {  \includegraphics[height= 3.4 cm]{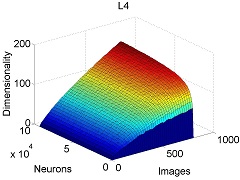} \label{Dimensionality3Da}}
\subfigure[] {  \includegraphics[height= 3.4 cm]{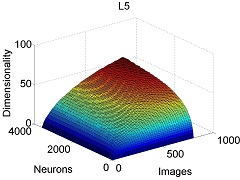} \label{Dimensionality3Db}}
\subfigure[] {  \includegraphics[height= 3.4 cm]{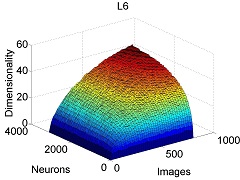} \label{Dimensionality3Dc}}
\subfigure[] {  \includegraphics[height= 3.4 cm]{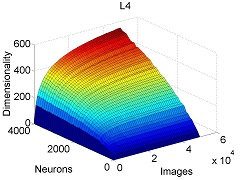} \label{Dimensionality3Dd}}
\subfigure[] {  \includegraphics[height= 3.4 cm]{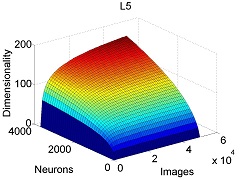} \label{Dimensionality3De}}
\subfigure[] {  \includegraphics[height= 3.4 cm]{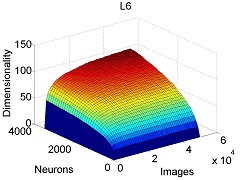} \label{Dimensionality3Df}}
\caption{Dimensionality surfaces corresponding to Layers $\{L4,L5,L6\}$: (a)(b)(c) Estimated dimensionality surfaces using the images from Dataset I; (d)(e)(f) Estimated dimensionality surfaces using the images from Dataset II.}\label{Dimensionality3D}
\end{figure}

When dealing with Dataset I, the number of images is sampled at increments of $20$ as done in \citep{Lehky2014}, i.e. the image subset sizes are $[20, 40, 60, . . . , 800]$. Considering
that the number of neurons at Layer $L4$ is much larger than those at Layers $L5$ and $L4$,
the number of neurons at Layer $L4$ is sampled at increments of $1000$, while
the number of neurons at Layer $\{L5,L6\}$ is sampled at increments of $20$ respectively.

When dealing with Dataset II, we compute the intrinsic dimensionality with the number of the neurons not more than $4000$ at Layer $L4$, since the occupied memories by the PCA-based method would exceed the hardware configuration in our PC if  all the images and all the neurons are considered. The number of neurons at the three layers is  sampled at increments of $100$, and the number of images is sampled at increments of $1000$.
Under each different combination of the image number and the neuron number, the sampling is independently done $10$ times, and the mean value of the $10$ estimated intrinsic dimensionality values is used as the final intrinsic dimensionality. Note that there exist a few neurons which
do not respond to any stimuli in our experiments, and they are simply removed before performing
the PCA-based method.

Figs. \ref{Dimensionality3Da}, \ref{Dimensionality3Db}, \ref{Dimensionality3Dc} show the
dimensionality surfaces corresponding to Layers $\{L4, L5, L6\}$ respectively on  Dataset I, and Figs. \ref{Dimensionality3Dd}, \ref{Dimensionality3De}, \ref{Dimensionality3Df} show the dimensionality surfaces corresponding to Layers $\{L4, L5, L6\}$ respectively on Dataset II. As seen from these figures, when the number of neurons and the number of images become large, the dimensionality surfaces vary smoothly in most cases.

The  two-step procedure for estimating the asymptotic dimensionality in Appendix B
 is implemented on Dataset I and Dataset II respectively.
There exist two fitting orders for computing the asymptotic dimensionality based on the one-dimensional asymptotic function (\ref{cfit}) in Appendix B: (I) ``neuron $\rightarrow$ image order": Firstly fitting along the number of neurons and then along the number of images;
(II) ``image $\rightarrow$ neuron order": Firstly fitting along the number of images and then along the  number of neurons. Both the two fitting orders are tested here, and Table \ref{largescale} lists the corresponding results on Dataset I and Dataset II.
As seen from this table, different layers in VGG have different asymptotic dimensionalities.
The higher a layer is, the smaller its asymptotic dimensionality is in most cases,
which implies the representation is more category-prototypical.
The asymptotic dimensionality for Dataset II at each referred layer is larger than that for Dataset I,
mainly because Dataset II is much larger than Dataset I.
These results confirm the existence of intrinsic dimensionality of
object representation around the order of hundred in \citep{Lehky2014}.


\begin{table}[t]
\caption{Estimated asymptotic dimensionality values in both the monkey inferotemporal cortex and the VGG network.}\label{largescale}
\footnotesize
\begin{center}
\begin{tabular}{|c|c|c|c|}
\hline
\multirow{2}{*}{Layer}     & \multirow{2}{*}{Fit order}   & \multicolumn{2}{c|}{Dimensionality}  \\
\cline{3-4}
          &      &  Dataset I &  Dataset II  \\
\hline
{Monkey inferotemporal cortex }      & neuron $\rightarrow$ image    &  87   &  -- \\
\hline
{Monkey inferotemporal cortex }      & image $\rightarrow$ neuron    &  105  &  -- \\
\hline
\hline
{VGG-L4}      & neuron $\rightarrow$ image    &  234   &  1504 \\
\hline
{VGG-L4}      & image $\rightarrow$ neuron    &  284   &  1569 \\
\hline
{VGG-L5}      & neuron $\rightarrow$ image    &  75    &  357 \\
\hline
{VGG-L5}      & image $\rightarrow$ neuron    &  102   &  336 \\
\hline
{VGG-L6}      & neuron $\rightarrow$ image    &  55    &  125 \\
\hline
{VGG-L6}      & image $\rightarrow$ neuron    &  74    &  117 \\
\hline

\end{tabular}
\end{center}
\end{table}

\paragraph{3.5.2 DNN Neurons Versus AIT Neurons on Intrinsic Dimensionality \\}

In \citep{Lehky2014}, it was concluded: (i) The intrinsic dimensionality
of object representations in primate AIT cortex was around 100; (ii) The asymptotic approximation process seems not
quite stable  due to
the limited number of data.

Our results show that, for both the small-sized Dataset I and the large-sized
Dataset II, the estimated dimensionality values by using the responses of Layers $\{L4,L5,L6\}$ of VGG
all tend to reach asymptotic limits, which supports the assumption that
the intrinsic dimensionality of object representation in primate AIT cortex reaches an asymptotic limit as both the number of stimulus images and the number of neurons approach to the infinity \citep{Lehky2014}.

In addition, it is noted from Table \ref{largescale} that for each referred layer of VGG,
the estimated dimensionality on Dataset II is larger than the estimated dimensionality on Dataset I as well as the estimated dimensionality (around 100) of object representation in primate IT cortex.
 It demonstrates that the estimated asymptotic dimensionality is fairly sensitive to the size of the testing dataset. Hence, a sufficiently
large dataset should be used for reliably estimating the asymptotic dimensionality of object representations in both DNNs and primate AIT cortex.
Furthermore, the estimated dimensionality on Dataset I at Layer $L4$ for object representation
 is  much larger than
that in primate IT cortex, indicating that
the estimated dimensionality in AIT cortex based on only less than $1000$ neurons might
not adequately reflect the intrinsic dimension of the object representation space,
and  a sufficiently
large number of sample neurons are necessary.



\section{Conclusion}

In this work, using the the concepts of single-neuron selectivity, population sparseness, and intrinsic dimensionality as introduced in \citep{Lehky2011,Lehky2014} for IT neural responses in monkey, the statistics of neural responses in DNNs are computed,
and here are some concluding points:
\begin{itemize}
\item The response statistics of the neurons at later layers of DNNs are not fully consistent with those of AIT neurons in monkey.
DNN neurons are more selective, and their population responses are not as
sparse as AIT neurons in monkey.

\item The changing trend of the response statistics from the low convolutional layers to the high fully-connected layers suggests that DNNs are able to firstly learn a large set of complex features for object representations through multiple convolutional layers, then learn a relatively small set of prototypical features for categorization through the fully-connected layers,
which could be the reason of why DNNs perform well on visual object categorization.
In addition, we thought object representation and categorization representation are two very different
issues, which should be distinguished.

\item The response statistics of DNN neurons with different combinations of
the neuron number and the stimulus  number might suggest that, the statistics of AIT neural responses
with only several hundreds of neurons and stimulus images might not be sufficient.
This is  because the statistics of DNN neural responses do not become stable until both the DNN neuron number and the image number reach a sufficiently large value (in our case, at least large than $10000$), although they are similar to those of AIT neurons when both the DNN neuron number and the image number are lower than $1000$.

\item The distribution of the intrinsic dimensionality of object representations in DNNs provides a support for the rationality that there exists an intrinsic dimensionality of object representation in primate AIT cortex \citep{Lehky2014}.
However, a sufficiently
large dataset and a sufficiently large number of sampled neurons should be used for reliably estimating the asymptotic dimensionality of object representations in both DNNs and primate AIT cortex.

\item IT neurons in primate are for both object representation and object categorization.
Object recognition requires distinguishing subtle differences among objects,
while object categorization requires removing differences within the same categorization.
Hence, we thought the object representation neurons should locate
anatomically anterior to those for object recognition and categorization.
We even wonder whether the AIT neurons recorded in \citep{Lehky2011}
are truly for object representations, not for object recognition or categorization, an issue
might be worth further elucidating in the future.

\end{itemize}

In sum, we seem the first to give a comparison on the response statistics of DNN neurons
with respect to those of AIT neurons in monkey.
The results could be of reference for both those people working
on the object recognition in primate and in DNNs.

\section*{Acknowledgements}
This
work was supported by the Strategic Priority Research Program of
the Chinese Academy of Sciences (XDB02070002). The authors would like
to thank Dr. Roozbeh Kiani for his provided data.

\section*{Appendix A: Kurtosis and Pareto Tail Index}

\paragraph{Kurtosis:}

Kurtosis (strictly speaking, excess kurtosis) is a measure of the ``peakedness'' of a probability distribution
for both single-neuron selectivity and population sparseness in many existing works \citep{Lehky2005,Lehky2007,Lehky2011,Tolhurst2009}. It only depends on the shape of the distribution, and is independent of the
mean or the variance.
Kurtosis  is defined as:
\begin{align}
Kurt = \frac{\frac{1}{N}\sum_{i=1}^N{(r_i - \bar{r}})^4}{[\frac{1}{N}\sum_{i=1}^N{(r_i - \bar{r})^2}]^2} - 3 \tag{A1} \label{Kurtosis}
\end{align}
where for single-neuron responses,
$r_i$ is the response of a neuron to
the $i$-th image, $N$ is the number of images; for population
responses, $r_i$ is the response of the $i$-th neuron to an
image,  $N$ is the number of neurons. $\bar{r} = \frac{1}{N}\sum_{i=1}^N{r_i}$ is the
mean response.

In addition, neurons in a population may have different activation levels in some cases,
then high population sparseness could arise as an artifact.
To alleviate this problem, the normalized data $r_i^n$,  which is obtained by dividing the response of each neuron by its mean response across all  the stimulus images, is also used for calculating
 kurtosis on both single-neuron selectivity and population sparseness:
\begin{align}
r_i^n =\frac{r_i}{\bar{r}} \tag{A2} \label{Knorm}
\end{align}
where $r_i$ is the response of a neuron to
the $i$-th image, $\bar{r} = \frac{1}{N}\sum_{i=1}^N{r_i}$ is the
mean response across all $N$ images.
According to (\ref{Kurtosis}) and (\ref{Knorm}), the normalization has no effect on single-neuron selectivity in principle,
but does have an effect on population sparseness.

\paragraph{Pareto Tail Index:}

The Pareto tail index \citep{Pickands1975} is utilized to analyse large responses occurring on the upper tails of
the probability density functions (PDFs). In \citep{Lehky2011}, tail data were fitted with
a generalized Pareto distribution by maximum likelihood.

The PDF for the generalized Pareto distribution is defined as:
\begin{align}
p=f(r|k, \sigma, \theta)=\frac{1}{\theta}(1 + k\frac{r - \theta}{\sigma})^{-1-\frac{1}{k}} \tag{A3} \label{Pareto}
\end{align}
where $\sigma$ is the scale, $\theta$
is the threshold where the upper tail of the probability density function starts, and $k$ is a shape
parameter quantifying heaviness of the tail, called the tail parameter.

Generally speaking,  if the kurtosis is large, it means the density function most probably has a heavy tail.
Similarly, if the tail index is large, the density function also has a heavy tail.
Since population sparseness and single-neuron selectivity are evaluated by the same criteria
 and computed by the same formulae, from the computational point of view, they are of no difference.

\section*{Appendix B: PCA-based Method for Intrinsic Dimensionality Estimation}

In \citep{Lehky2014}, based on  the recorded neural responses to a set of stimulus images,
the PCA-based method computed the intrinsic dimensionality of object
representations  by the following  two steps:

(1) Different subsets with different combinations of the image number and the neuron number were sampled from the original dataset. For each subset with a given number of neurons and images as well as a randomly reshuffled version of it, PCA  was performed on them respectively.
Each set of eigenvalues was normalized to sum to 1, and was sorted in a descending manner respectively.
The eigenvalues of the original data which were larger than those of the corresponding reshuffled data were considered to  reflect the signals in the original data, while the rest were considered noise. The number of the large eigenvalues of reflecting the signals was considered as the intrinsic dimensionality of the original data.
The authors \citep{Lehky2014} observed that for a given set of data, only one reshuffling was sufficient,
and there was little change in the intrinsic dimensionality estimation for repeated reshufflings.

(2) Assuming that the intrinsic dimensionality reached an asymptotic limit as both the image number and the neuron number approached to infinity, an asymptotic dimensionality was computed with the obtained dimensionality values on different subsets as the final dimensionality. The asymptotic dimensionality was calculated in a two-step process: a one-dimensional asymptotic
function was firstly fit along one parameter (either number of neurons or
number of images), then by fixing the first parameter estimation,
the asymptotic approximation along the other parameter was carried out.
Lehky et al. \citep{Lehky2014} used two curve-fitting functions respectively for estimating asymptotic dimensionality, and found that the estimated dimensionality values by the two functions were quite close. Hence,
we only use the following fitting function (the function (2.4) in \citep{Lehky2014})  for estimating
the asymptotic dimensionality here:
\begin{align}
z = a[1-(\frac{b}{\mathrm{exp}(\frac{x^c}{d})-1+b})^e]^f \tag{A4} \label{cfit}
\end{align}
where $z$ was the dimensionality, $x$ was the number of images or neurons,
and $\{a,b,c,d,e,f\}$ were fitting parameters.


 Noting that all nonlinear curve-fitting algorithms were dependent on parameter initials, it was hard to give an appropriate set of initials for (\ref{cfit}) manually. Hence,
each curve was repeatedly fit by trial-and-error setting of initial parameters
until the fitting error was lower than a predefined threshold $\varepsilon$.




\end{document}